\documentclass{Interspeech2024}

\interspeechcameraready

\title{Comparing Discrete and Continuous Space LLMs for Speech Recognition}

\usepackage{graphicx}
\usepackage{subcaption}
\usepackage{multirow}

\name[affiliation={1}]{Yaoxun}{Xu}
\name[affiliation={2}]{Shi-Xiong}{Zhang}
\name[affiliation={2}]{Jianwei}{Yu}
\name[affiliation={1,3,\dagger}]{Zhiyong}{Wu}
\name[affiliation={2}]{Dong}{Yu}

\address{
  $^1$ Shenzhen International Graduate School, Tsinghua University, Shenzhen, China\\
  $^2$ Tencent AI Lab\\
  $^3$ The Chinese University of Hong Kong, Hong Kong SAR, China}
\email{xuyx22$@$mails.tsinghua.edu.cn, zhangshixiong$@$gmail.com, \\tomasyu$@$tencent.com, zywu$@$sz.tsinghua.edu.cn\thanks{$\dagger$ Corresponding author.}}


\begin{document}

\maketitle

\begin{abstract}
This paper investigates discrete and continuous speech representations in Large Language Model (LLM)-based Automatic Speech Recognition (ASR), organizing them by feature continuity and training approach into four categories: supervised and unsupervised for both discrete and continuous types. We further classify LLMs based on their input and autoregressive feedback into continuous and discrete-space models. Using specialized encoders and comparative analysis with a Joint-Training-From-Scratch Language Model (JTFS LM) and pre-trained LLaMA2-7b, we provide a detailed examination of their effectiveness. Our work marks the first extensive comparison of speech representations in LLM-based ASR and explores various modeling techniques. We present an open-sourced achievement of a state-of-the-art Word Error Rate (WER) of 1.69\% on LibriSpeech using a HuBERT encoder, offering valuable insights for advancing ASR and natural language processing (NLP) research.
    
\end{abstract}

\section{Introduction}

Automatic Speech Recognition (ASR) \cite{asr00,asr01} is pivotal in speech processing and human-machine interaction. A critical aspect of ASR is the representation of raw speech signals, where redundancy poses a challenge to computational efficiency. This challenge has spurred the development of more compact speech representations, divided into discrete and continuous forms. While discrete representations offer cost efficiency with limited information capacity, continuous representations, despite being more expensive, encapsulate a richer array of information.

Traditional ASR models \cite{gales2008application, asr0} utilize discrete speech units, like phonemes \cite{yusnita2011phoneme} or triphones \cite{chen2014joint}, for continuous speech feature modeling, encompassing components such as acoustic models (AMs), pronunciation lexicons and language models (LMs). These units serve as discrete intermediate representations of continuous speech signals in these cascaded systems.
Alternatively, end-to-end ASR frameworks \cite{graves2012sequence, chan2015listen, Whisper}, jointly modeling the continuous acoustic signal and discrete language models, have gradually become mainstream. These frameworks no longer treat ASR systems as cascaded systems (AM and LM) but instead, they directly transform speech features into continuous representations through encoder \cite{gulati2020conformer, yao2023zipformer,gao2022paraformer} and joint model them with word embeddings in the continuous space.

In the realm of natural language processing (NLP), language models conventionally employ discrete text tokens as processing units. Recently, the rise of large language models (LLMs) has notably advanced NLP \cite{llm,llm1}, thanks to increased data availability and computational power. Applying LLMs' advanced language understanding and generation to improve ASR has become a key research area. 
For instance, VioLA \cite{wang2023viola} converts audio signals into discrete codecs via a pretrained codebook, then autoregressively generates and decodes tokens to text. 
SpeechGPT \cite{zhang2023speechgpt} and AudioPaLM \cite{rubenstein2023audiopalm} use K-means to convert continuous audio into discrete forms for processing with pretrained LLaMA \cite{touvron2023LLaMA} and PaLM  \cite{anil2023palm} models, respectively. 
These methods incorporate discretized speech tokens into LLMs, while others integrate continuous speech features. \cite{cite} feeds HuBERTCTC and HuBERT's continuous outputs to GPT-NeoX \cite{neox}, directly leverages continuous speech features. Models like SALM \cite{chen2023salm}, Whispering LLaMA \cite{radhakrishnan2023Whispering}, SALMONN \cite{tang2023salmonn}, and Qwen-Audio \cite{chu2023qwen} also integrate continuous speech with LLMs via adapters, showcasing diverse methods to combining ASR with advanced language modeling.


\begin{figure*}[th]
\centering
\includegraphics[width=2.0\columnwidth]{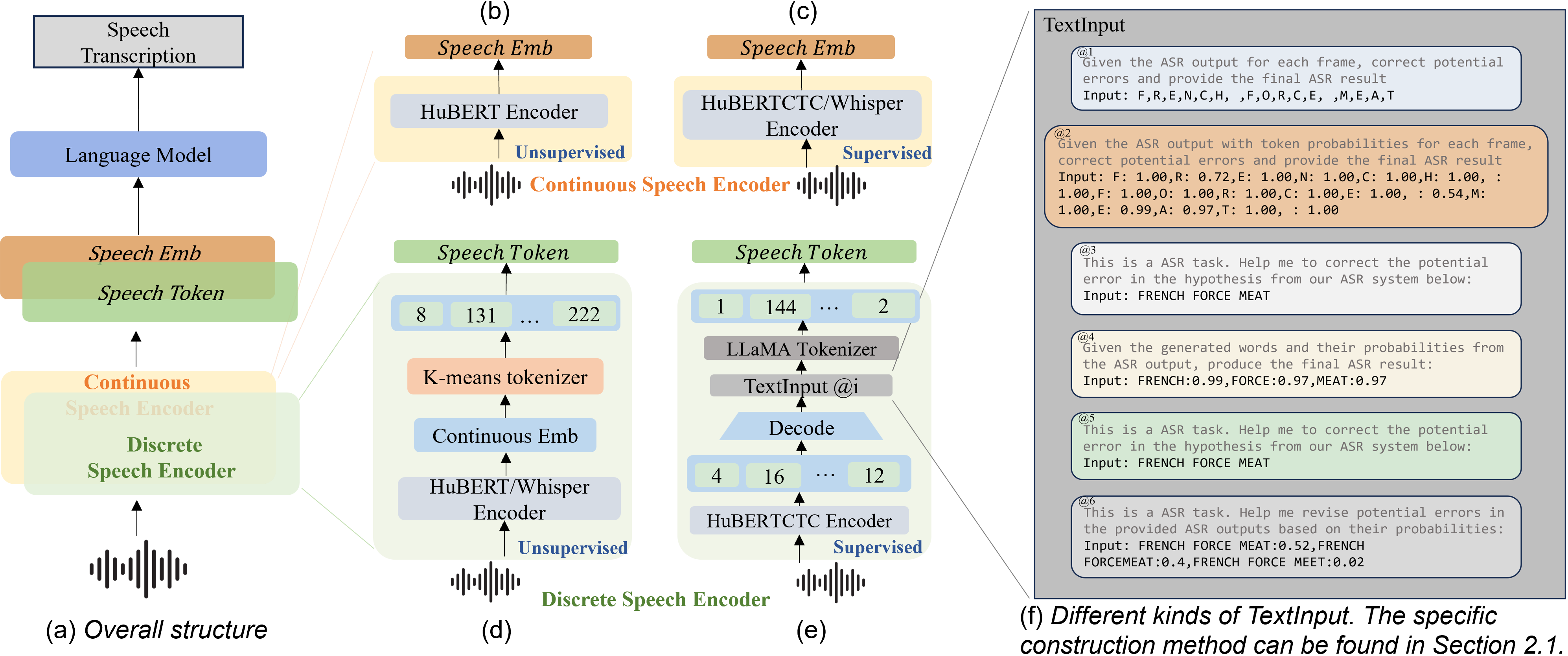} 

\caption{Overall architectures for continuous and discrete speech encoders. Figures (b) to (e) illustrate four distinct speech encoders, each extracting a different type of speech representation.}
\vspace{-0.5cm}

\label{fig1}
\end{figure*}


Despite the advancements in LLM-based speech recognition, there's a noticeable gap in systematically analyzing discrete versus continuous speech representations within this domain. Our study seeks to fill this void by exploring these representations in LLM-based speech recognition tasks, offering a detailed comparison to guide future research. We classify speech into discrete and continuous categories based on the nature of speech features and further distinguish them by whether they utilize paired speech-transcription data in training the speech encoder, leading to four distinct groups: supervised and unsupervised discrete speech representations, along with supervised and unsupervised continuous speech representations. For each category, we create specialized speech encoder and models to conduct comprehensive comparisons using both a Joint-Training-From-Scratch Language Model (JTFS LM) and the pretrained LLaMA2-7b \cite{touvron2023LLaMA2} as benchmarks, shedding light on the effectiveness of these four types of speech representations.
The main contributions of this study are as follows:
\begin{itemize}
\item To the best of our knowledge, this is the first comprehensive comparative study focusing on discrete and continuous speech representations in LLM-based speech recognition.

\item This study proposes and evaluates different modeling approaches within discrete and continuous space LLMs.

\item To our knowledge, this represents the state-of-the-art in open-sourced models on LibriSpeech, achieving a WER of 1.69\%.\footnote{The code can be found: https://github.com/xuyaoxun/ASRCompare}

\end{itemize}

\section{LLM based Speech Recogntion}

As illustrated in Figure 1(a), we utilize discrete or continuous speech encoders to preprocess speech, yielding either discrete tokens or continuous embeddings. These are then integrated into language models to produce the final transcription results.

\subsection{Discrete V.S. Continuous Speech Representations}

For the four distinct types of speech representations, we pinpoint and select highly representative features within each category, subsequently designing corresponding feature extractors.

\noindent\textbf{Continuous Unsupervised Representation} 

In Figure 1(b), we use the pre-trained HuBERT \cite{hsu2021hubert} model as speech encoder. HuBERT is a self-supervised learning model designed for speech processing tasks. To evaluate its layer-wise extraction capabilities, we select representations from the 0th layer (post-preprocessing and before entering the HuBERT encoder) and the 8th, 16th, and 24th layers of HuBERT-large as continuous unsupervised speech representations.

\noindent\textbf{Continuous Supervised Representation} 

In Figure 1(c), we use the pre-trained HuBERT-CTC model as speech encoder. This model augments HuBERT with a prediction layer and reduces dimensions to 32 modeling units, utilizing additional speech-text pairs and CTC \cite{ctc} loss for optimization. We employ the 24th layer of HuBERT and the 32-dim prediction features from the projection layer. Moreover, we adopt Whisper's \cite{Whisper} encoder as an alternative extractor, which is an encoder-decoder model trained on numerous speech-text pairs, yielding another continuous supervised representation.

\noindent\textbf{Discrete Unsupervised Representation} 

In Figure 1(d), we train K-means clustering extractors with 500, 1000, and 1500 clusters on a subset of training dataset. Using these K-means extractors, we categorize continuous speech representations and remove duplicates to form discrete unsupervised speech representations.

\noindent\textbf{Discrete Supervised Representation} 

In Figure 1(e), we obtain high-probability tokens from pre-trained HuBERT-CTC logits using softmax, map them to text, and apply post-processing strategies for unique textual prompts.
As shown in Figure 1(f), method @1 maps deduplicated HuBERT-CTC logits into the text domain, separating characters with commas. Method @2 adds character probabilities to the input. Method @3 forms a sentence by concatenating characters. Method @4 calculates the mean probability of tokens in a word, based on HuBERT's original logits, and appends it to the input. Method @5 uses the highest-scoring sentence after rescoring speech logits via HuBERT-CTC with a pre-trained 4-gram language model. Method @6 rescores HuBERT-CTC logits using a pre-trained 4-gram language model, providing the top three results and their probabilities as input. After obtaining the TextInput, we seamlessly integrate it by passing it through the LLaMA2 tokenizer to generate discrete tokens, which serve as the supervised discrete representation.
\vspace{-0.3cm}
\subsection{Discrete V.S. Continuous Modeling}
\vspace{-0.1cm}
\begin{figure*}[!htbp]
  \centering
  \begin{subfigure}[b]{0.39\textwidth}
        \includegraphics[width=\textwidth]{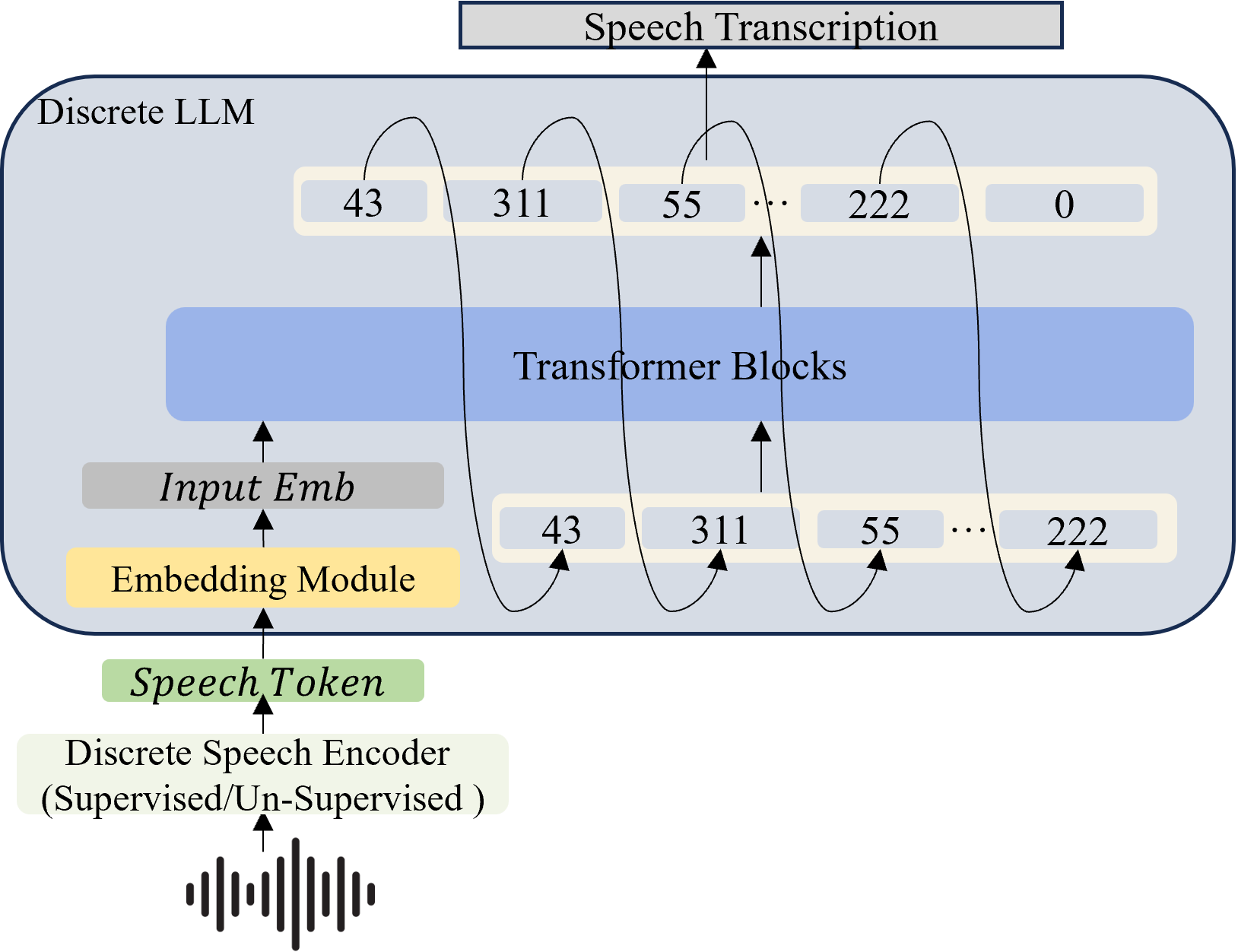}
        \caption{Model design for discrete scenarios}
    \end{subfigure}
    \hspace{1.5cm}
    \begin{subfigure}[b]{0.40\textwidth}
        \includegraphics[width=\textwidth]{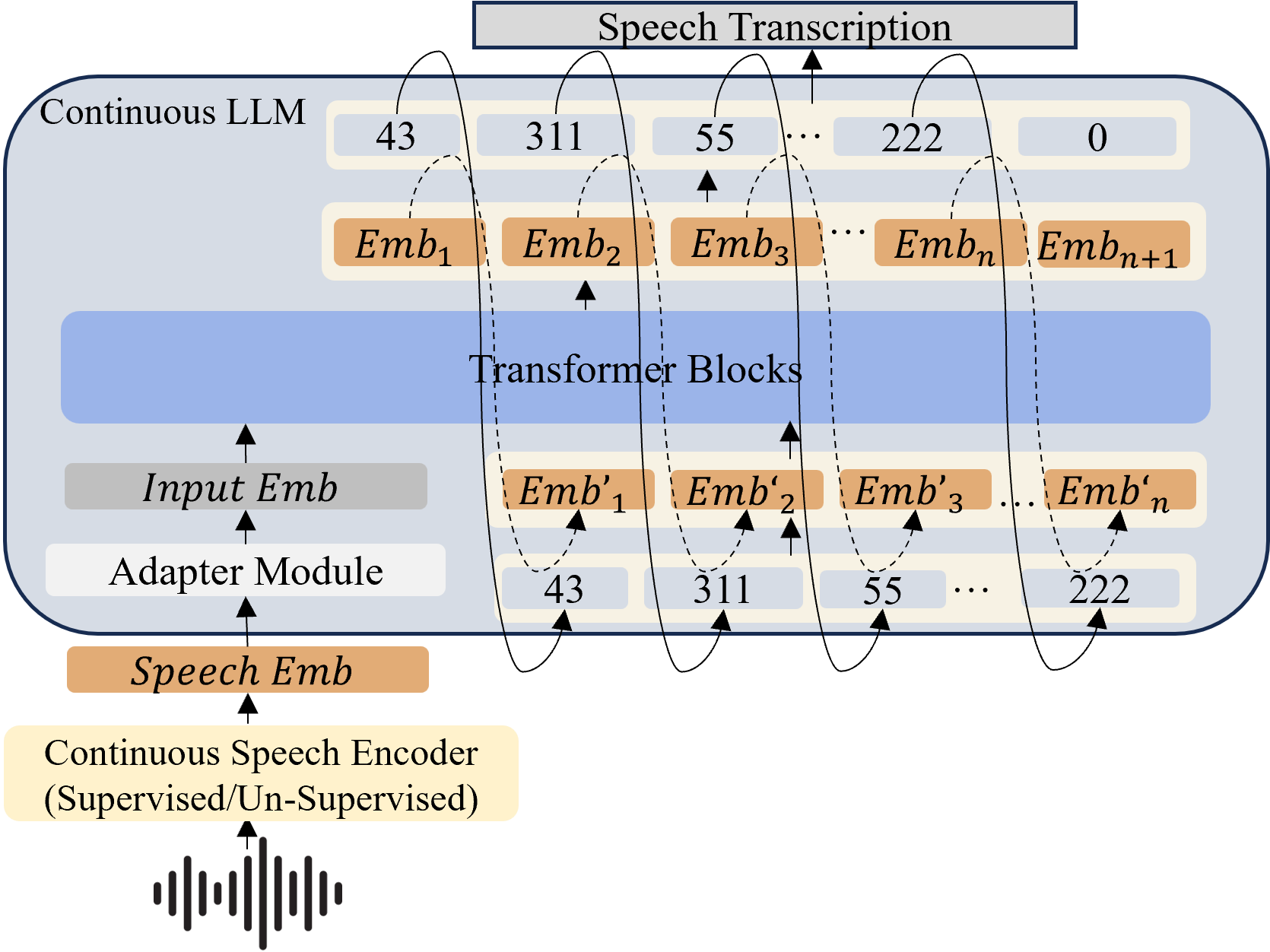}
        \caption{Model design for continuous scenarios}
    \end{subfigure}
    \vspace{-0.2cm}
    \caption{Model design for discrete and continuous scenarios. In Figure 2(b), dashed lines show the data flow for the JTFS LM, and solid lines for the LLaMA2 model}
    \vspace{-0.4cm}
\end{figure*}
To comprehensively compare discrete and continuous speech representations, we devise models tailored specifically for each type of representation, as illustrated in Figure 2. Specifically, we have employed the Transformer \cite{vaswani2017attention}-based language models, LLaMA2 and JTFS LM, as benchmarks for comparison. 

In the discrete scenario shown in Figure 2(a), we employ discrete speech encoders (detailed in Section 2.1) to convert speech segments into speech tokens. These tokens are passed through an embedding module to obtain input embeddings, which are then fed into the Transformer blocks. For discrete supervised speech features, the embedding module consists of the text embedding layer from the LLaMA2 tokenizer. In contrast, for discrete unsupervised speech representations, the corresponding embedding module comprises an embedding layer mapping from the K-means cluster count to the Transformer dimensions, along with a two-layer perceptron. During this phase, the JTFS LM trains all parameters jointly, while LLaMA2 is fine-tuned using the LoRA \cite{hu2021lora} method. Text sequences are generated sequentially through an autoregressive approach.

In the continuous scenario depicted in Figure 2(b), we extract continuous speech embeddings from the continuous speech encoder (as detailed in Section 2.1). These embeddings are then processed through an adapter module composed of a two-layer perceptron to obtain the input embeddings. The input embeddings are fed into the Transformer blocks to generate text sequences autoregressively. Notably, for the JTFS LM (dashed line in Figure 2(b)), we avoid discretizing the Transformer blocks' output to prevent information loss. Instead, we directly transfer the output to the input side, preserving its integrity. Concurrently, the Transformer blocks' output is projected through a mapping layer to obtain the token sequence in the dictionary, continuing until a termination symbol is reached. For LLaMA2, similar to the discrete scenario, we discretize the Transformer blocks' output and pass it through the LLaMA2 tokenizer before re-inserting it into the Transformer blocks.

\subsection{Training Loss}
In the case of discrete scenarios (Figure 2(a)) and continuous scenarios with LLaMA2 (solid lines in Figure 2(b)), we employ the cross-entropy loss (CELoss) as the loss function during training:
\[
\text{loss} = \text{CELoss}(\textit{target}, \textit{predict})
\]
Here, $\textit{target}$ denotes the ground truth labels, and $\textit{predict}$ represents the predicted output.

For the continuous LLM with JTFS LM (dashed lines in Figure 2(b)), we employ MSELoss alongside CELoss, ensuring consistency between the Transformer Blocks' input and output while enabling autoregressive generation:
\[
\text{loss} = \text{CELoss}(\textit{target}, \textit{predict}) + \alpha * \text{MSELoss}(\textit{emb\_in}, \textit{emb\_out})
\]
In this equation, $\alpha$ is a hyperparameter that controls the balance between the two loss components. $\textit{emb\_in}$ refers to the input embedding, while $\textit{emb\_out}$ denotes the output embedding.

\vspace{-0.2cm}
\section{Experiment}
We perform experiments on the LibriSpeech \cite{panayotov2015librispeech} dataset, comprising 960.9 hours of training, 10.7 hours of validation, 5.4 hours of test-clean, and 5.1 hours of test-other audio. We evaluate performance using WER on test-clean and test-other sets.
Our JTFS LM consists of 10 stacked Transformer layers with 1024 dimensions and 8 attention heads. In discrete speech representation, we create distinct embedding layers and two-layer perceptrons to project discrete representations into the Transformer's space, depending on the number of clustered tokens. For continuous representations, we use a two-layer perceptron to project speech representations into the Transformer's or LLaMA2's space. In supervised discrete speech representation, we fine-tune LLaMA with LoRA, using rank 16, alpha 16, target modules for gate projection, down projection, and up projection layers, and a dropout rate of 0.05.
We set the learning rate at 1e-5, $\alpha$ at 100, and train the model using 8 A100 GPUs.
\vspace{-0.2cm}
\section{Result}
\subsection{Results on joint-training-from-scratch language model}

We conduct an in-depth analysis of the experimental results derived from the JTFS LM, examining the effects of various configurations. These include the type of speech representation (discrete vs. continuous), the choice of speech encoder (HuBERT vs. Whisper), the number of K-means clusters, and the layer of the HuBERT model.

\begin{table}[!h]
\vspace{-0.1cm}
\caption{Experiment results on JTFS LM} \vspace{-0.2cm}
\scalebox{0.83}{

\begin{tabular}{cccccc}
\toprule[2pt]
Speech Type                                                                        & Encoder                  & \begin{tabular}[c]{@{}c@{}}Kmeans\\ Cluster\end{tabular} & Layer & ID   & \begin{tabular}[c]{@{}c@{}}WER(\%)\\ clean/other\end{tabular} \\ \hline
\multirow{15}{*}{\begin{tabular}[c]{@{}c@{}}Discrete\\ Unsupervised\end{tabular}}  & \multirow{12}{*}{HuBERT} & \multirow{4}{*}{500}                                     & 0     & \#1  & 109.47/109.54                                             \\
                                                                                   &                          &                                                          & 8     & \#2  & 96.62/98.13                                               \\
                                                                                   &                          &                                                          & 16    & \#3  & 80.54/81.79                                               \\
                                                                                   &                          &                                                          & 24    & \#4  & 76.8/77.51                                                \\ \cline{3-6} 
                                                                                   &                          & \multirow{4}{*}{1000}                                    & 0     & \#5  & 106.04/107.13                                             \\
                                                                                   &                          &                                                          & 8     & \#6  & 89.12/93.8                                                \\
                                                                                   &                          &                                                          & 16    & \#7  & 72.2/75.57                                                \\
                                                                                   &                          &                                                          & 24    & \#8  & 74.08/72.69                                               \\ \cline{3-6} 
                                                                                   &                          & \multirow{4}{*}{1500}                                    & 0     & \#9  & 119.01/117.2                                              \\
                                                                                   &                          &                                                          & 8     & \#10 & 87.33/92.11                                               \\
                                                                                   &                          &                                                          & 16    & \#11 & 70.13/71.91                                               \\
                                                                                   &                          &                                                          & 24    & \#12 & 71.85/70.39                                               \\ \cline{2-6} 
                                                                                   & \multirow{3}{*}{Whisper} & 500                                                      & \textemdash     & \#13 & 75.59/76.94                                               \\
                                                                                   &                          & 1000                                                     & \textemdash     & \#14 & 58.27/60.01                                               \\
                                                                                   &                          & 1500                                                     & \textemdash     & \#15 & 49.18/51.90                                               \\ \hline
\multirow{4}{*}{\begin{tabular}[c]{@{}c@{}}Continuous\\ Unsupervised\end{tabular}} & \multirow{4}{*}{HuBERT}  & \textemdash                                                        & 0     & \#16 & 41.13/67.55                                               \\
                                                                                   &                          & \textemdash                                                        & 8     & \#17 & 12.47/23.93                                               \\
                                                                                   &                          & \textemdash                                                        & 16    & \#18 & 7.17/12.21                                                \\
                                                                                   &                          & \textemdash                                                        & 24    & \#19 & 18.27/26.14                                               \\ \hline
\begin{tabular}[c]{@{}c@{}}Continuous\\ Supervised\end{tabular}                    & Whisper                  & \textemdash                                                        & \textemdash     & \#20 & 5.28/9.74  \\
\bottomrule[2pt]
\end{tabular}

}
\vspace{-0.2cm}

\end{table}

\noindent\textbf{The Impact of Continuous and Discrete Configurations:} The continuous configuration outperforms the discrete one, as shown by the WER comparisons between experiments \#11 and \#18, and \#15 and \#20. This notable reduction in WER in the continuous setting highlights its effectiveness. The advantage stems from the Joint-Training-From-Scratch Language Model (JTFS LM), which, beginning with random initialization, leverages either discrete or continuous speech representations. Discrete tokens undergo significant information loss during clustering, whereas continuous representations retain most of the information crucial for ASR. This indicates that information loss increases with the discreteness of tokens.

\noindent\textbf{The Impact of Different Encoder: }
 The choice of the encoder is critical, with Whisper encoders consistently outperforming HuBERT encoders in both discrete (with identical K-means) and continuous settings, as shown in comparisons \#12 with \#15, and \#18 with \#20. These results highlight Whisper's enhanced feature extraction ability, indicating that supervised training can further improve the alignment of speech representations to text, thus boosting performance.

\noindent\textbf{The Impact of K-means Clusters Number:}
WER consistently decreases with an increase in the number of K-means clusters, evident from comparisons \#13, \#14, and \#15. This pattern holds true for the HuBERT encoder, suggesting that more clusters enhance performance by capturing a greater volume of information. However, a higher cluster count also means a more complex feature extraction process, which may demand additional computational resources and time. Therefore, balancing the number of clusters with resource constraints is key to optimizing performance.

\noindent\textbf{The Impact of Encoder Layers:}
In the continuous setting using the HuBERT encoder, there's a marked decrease in WER from layer 0th to 16th. However, an uptick in WER is observed at layer 24th, as evidenced by \#16 showing higher WER compared to \#17 and \#18, with \#19 experiencing a slight increase. This could be due to HuBERT's focus on predicting cluster IDs during training, potentially leading to the capture of more acoustically driven but semantically irrelevant features. 

\subsection{Results on LLaMA2} 
\vspace{-0.3cm}

\begin{table}[h]
\caption{LLaMA2 Exp. For details of TextInput@i see Fig.~\ref{fig1}. HuBERT24 Emb refers to HuBERT's 24th layer features, and HuBERTCTC Emb refers to HuBERTCTC's final layer features.}\vspace{-0.2cm}
\scalebox{0.83}{



\begin{tabular}{cccc}
\toprule[2pt]
Speech Type                                                                      & Method                                                                     & ID   & \begin{tabular}[c]{@{}c@{}}WER(\%)\\ clean/other\end{tabular} \\ \hline
\multirow{9}{*}{\begin{tabular}[c]{@{}c@{}}Discrete\\ Supervised\end{tabular}}   & HuBERTCTC                                                                     & \#21 & 2.08/4.24                                                 \\
                                                                                 & HuBERTCTC+4-gram                                                              & \#22 & 1.82/3.59                                                 \\
                                                                                 & TextInput@1+LLaMA2                                                         & \#23 & 2.14/4.13                                                 \\
                                                                                 & TextInput@2+LLaMA2                                                         & \#24 & 2.52/4.41                                                 \\
                                                                                 & TextInput@3+LLaMA2                                                         & \#25 & 1.99/4.03                                                 \\
                                                                                 & TextInput@4+LLaMA2                                                         & \#26 & 1.96/3.97                                                 \\
                                                                                 & TextInput@5+LLaMA2                                                         & \#27 & 1.72/3.57                                                 \\
                                                                                 & TextInput@6+LLaMA2                                                         & \#28 & 1.80/3.61                                                 \\
                                                                                 & \begin{tabular}[c]{@{}c@{}}HuBERTCTC(xlarge)+\\ 4-gram+LLaMA2\end{tabular} & \#29 & \textbf{1.69/3.03}                                                 \\ \hline
\begin{tabular}[c]{@{}c@{}}Discrete\\ Unsupervised\end{tabular}                  & HuBERT24-1500+LLaMA2                                                       & \#30 & 65.26/75.85                                               \\ \hline
\begin{tabular}[c]{@{}c@{}}Continuous\\ Unsupervised\end{tabular}                & HuBERT24 Emb+LLaMA2                                                        & \#31 & 13.05/16.77                                               \\ \hline
\multirow{2}{*}{\begin{tabular}[c]{@{}c@{}}Continuous\\ Supervised\end{tabular}} & HuBERTCTC24 Emb+LLaMA2                                                     & \#32 & 6.26/7.09                                                 \\
                                                                                 & HuBERTCTC Emb+LLaMA2                                                       & \#33 & 9.99/11.93    \\
\bottomrule[2pt]
\end{tabular}

}

\end{table}


\noindent\textbf{LLM as Discrete Error Token Corrector:}
In discrete supervised scenarios, encompassing Experiments \#21 to \#29, LLMs function as correctors of erroneous tokens. Here, the encoder initially converts continuous speech into discrete tokens, which are then refined by LLMs. This correction relies on long-context LM probabilities, effectively acting as a second-pass decoding. For instance, the comparison between \#22 and \#21 demonstrates how employing a pre-trained 4-gram language model in the first-pass WFST decoding \cite{miao2015eesen} significantly improves performance by leveraging enhanced contextual information. Moreover, LLMs further reduce WER, as seen in comparisons \#27 with \#22.
Comparing models \#23 with \#25 and \#24 with \#26 shows LLaMA2's enhanced word sensitivity and error correction, surpassing character-level input.


\noindent\textbf{N-best and Confidence Score:}
When comparing \#28 with \#27 and \#22, it's observed that although \#28, offering 3-best results, enriches the information set, it doesn't outperform \#27, yet it does exceed \#22 in performance. A comparison between \#26 and \#25 shows that \#26, with additional confidence scores, achieves marginally better results. This suggests that while LLaMA2 benefits from extra data, optimizing its 7B parameters for improved outcomes might necessitate more comprehensive N-best lists and confidence score inputs.

\noindent\textbf{State-of-The-Art:}
Using the TextInput@5 approach with an upgrade from the HuBERT-large to the xlarge encoder, we achieved a WER of 1.6/3.0, as demonstrated in \#29. This marks, to our knowledge, the best-reported WER on LibriSpeech using a HuBERT encoder. Our performance surpasses that of \cite{hsu2021hubert}, which reported a WER of 1.8\% using the same HuBERT encoder. While \cite{zhang2020pushing} reached a lower WER of 1.5\%, their model uses more data and has not been made publicly available.


\noindent\textbf{Supervised \textit{V.S.} Unsupervised:}
Model \#27, which employs HuBERT-CTC trained with transcribed data to generate discrete tokens, significantly outperforms Model \#30. The latter relies on an unsupervised clustering algorithm for token production. This performance disparity is credited to LLaMA2's advanced sensitivity to textual features, whereas K-means method used in the unsupervised approach focuses more on auditory characteristics, offering a narrower grasp of semantic content. Additionally, the constrained cluster count in unsupervised learning causes phonetically similar sounds to be merged into the same category, diminishing the model's ability to distinguish nuanced pronunciation differences, thus impacting recognition accuracy.

In continuous methods, supervised Model \#32 outshines unsupervised Model \#31 due to fine-tuning with speech-text pairs, resulting in text-oriented representations and lower WER, showcasing the advantages of supervised learning in achieving nuanced speech recognition.

\noindent\textbf{Impact of Matched Tokens:}
Comparing Discrete Supervised with Continuous Supervised methods reveals that despite the richer data from continuous representations, \#27 still outperforms \#33. This advantage is due to \#27 generating token sequences that matched with LLaMA2's pretraining tokens, enhancing compatibility and performance.

\noindent\textbf{Continuous Unsupervised \textit{V.S.} Discrete Unsupervised:}
In the comparison between Discrete Unsupervised and Continuous Unsupervised methods, \#31 surpasses \#30. The disparity in performance stems from the discrete tokens derived via clustering, focusing on frame-level acoustic features. The constrained cluster categories in discrete representations lead to a significant loss of acoustic details. As a result, discrete representations, being less informative than their continuous counterparts, present a more substantial challenge for the language model in recognizing speech accurately.

\noindent\textbf{JTFS LM \textit{V.S.} LLM:}
A comparison of \#31 with \#19 and \#30 with \#12 clearly reveals that the pre-trained LLaMA2 model delivers superior recognition results for both discrete unsupervised and continuous unsupervised representations. This outcome underscores the effectiveness of pre-training in enhancing language model performance. The pre-trained LLaMA2 model utilizes its prior knowledge, gleaned from extensive text data, to effectively recognize and decode speech representations, regardless of whether they are discrete or continuous.

\section{Conclusion}

This study investigates discrete and continuous speech representations in LLM-based ASR, organizing them into supervised and unsupervised categories for both types. We further classify LLMs into continuous and discrete-space models according to their input and feedback mechanisms. Through the development of specialized speech encoders and detailed comparative analysis with a Joint-Training-From-Scratch Language Model and the pre-trained LLaMA2, we conduct the first thorough assessment of these representations' impact on LLM-based ASR. Our rigorous experiments have led to a state-of-the-art, open-sourced WER of 1.69\% on the LibriSpeech dataset.
\section{Acknowledgement}
This work is supported by National Natural Science Foundation of China (62076144), Shenzhen Science and Technology Program (WDZC20220816140515001, JCYJ20220818101014030) and Tencent AI Lab Rhino-Bird Focused Research Program (RBFR2023015).
\bibliographystyle{IEEEtran}
\bibliography{mybib}

\end{document}